\documentclass{article}

\usepackage[preprint,nonatbib]{neurips_2022}

\usepackage[utf8]{inputenc} 
\usepackage[T1]{fontenc}    
\usepackage{hyperref}       
\usepackage{url}            
\usepackage{booktabs}       
\usepackage{amsfonts}       
\usepackage{nicefrac}       
\usepackage{microtype}      
\usepackage{xcolor}         

\usepackage{algorithmic}
\usepackage[ruled,vlined]{algorithm2e}
\usepackage{graphicx}
\usepackage{amssymb}
\usepackage{mathtools}
\usepackage{amsmath}
\usepackage{comment}
\usepackage{caption} 
\usepackage{svg}
\usepackage{subcaption}
\usepackage{amsthm}
\theoremstyle{plain}
\newtheorem{assumption}{Assumption}

\newtheorem{definition}{Definition}
\newtheorem{remark}{Remark}
\newtheorem{theorem}{Theorem}
\newtheorem{lemma}{Lemma}

\newenvironment{customthm}[1]
  {\renewcommand\thetheorem{#1}\theorem}
  {\endtheorem}

\newcommand{\squishlist}{
 \begin{list}{$\bullet$}
  { \setlength{\itemsep}{0pt}
     \setlength{\parsep}{5pt}
     \setlength{\topsep}{5pt}
     \setlength{\partopsep}{0pt}
     \setlength{\leftmargin}{1.5em}
     \setlength{\labelwidth}{1em}
     \setlength{\labelsep}{0.5em} } }
 \newcommand{\squishend}{\end{list}}

\usepackage{mathtools}

\title{Convergence Analysis of Decentralized ASGD}

\author{%
  Mauro DL Tosi \\
  Department of Computer Science\\
  University of Luxembourg\\
  Esch-sur-Alzette, Luxembourg \\
  \texttt{mauro.dalleluccatosi@uni.lu} \\
  \And
  Martin Theobald \\
  Department of Computer Science\\
  University of Luxembourg\\
  Esch-sur-Alzette, Luxembourg \\
  \texttt{martin.theobald@uni.lu} \\
}

\begin{document}
\maketitle
\begin{abstract}
Over the last decades, Stochastic Gradient Descent (SGD) has been intensively studied by the Machine Learning community. Despite its versatility and excellent performance, the optimization of large models via SGD still is a time-consuming task. To reduce training time, it is common to distribute the training process across multiple devices. Recently, it has been shown that the convergence of {\em asynchronous SGD} (ASGD) will always be faster than mini-batch SGD. However, despite these improvements in the theoretical bounds, most ASGD convergence-rate proofs still rely on a centralized parameter server, which is prone to become a bottleneck when scaling out the gradient computations across many distributed processes.

In this paper, we present a novel convergence-rate analysis for {\em decentralized and asynchronous SGD} (DASGD) which does not require partial synchronization among nodes nor restrictive network topologies. Specifically, we provide a bound of $\mathcal{O}(\sigma\epsilon^{-2}) + \mathcal{O}(QS_{avg}\epsilon^{-3/2}) + \mathcal{O}(S_{avg}\epsilon^{-1})$ for the convergence rate of DASGD, where $S_{avg}$ is the {\em average staleness} between models, $Q$ is a constant that bounds the norm of the gradients, and $\epsilon$ is a (small) error that is allowed within the bound. Furthermore, when gradients are not bounded, we prove the convergence rate of DASGD to be $\mathcal{O}(\sigma\epsilon^{-2}) + \mathcal{O}(\sqrt{\hat{S}_{avg}\hat{S}_{max}}\epsilon^{-1})$, with $\hat{S}_{max}$ and $\hat{S}_{avg}$ representing a loose version of the {\em average} and {\em maximum staleness}, respectively. Our convergence proof holds for a fixed stepsize and any non-convex, homogeneous, and L-smooth objective function. We anticipate that our results will be of high relevance for the adoption of DASGD by a broad community of researchers and developers.
\end{abstract}

\section{Introduction}
Over the last decades, Stochastic Gradient Descent (SGD) \cite{robbins1951stochastic} has been intensively studied by the Machine Learning community. SGD and its many variants (mini-batch SGD \cite{lecun2002efficient}, ADAM \cite{kingma2014adam}, AdaGrad \cite{duchi2011adaptive}, etc.) have demonstrated their robustness by frequently achieving state-of-the-art results in diverse problems. In particular, Large Language Models (LLMs) such as GPT-4 \cite{openai2023gpt4}, Generative Models such as Stable Diffusion \cite{rombach2021highresolution}, and other Neural Network (NN) models such as \cite{chen2023symbolic, schafl2022hopular, zhang2020pushing}, involving billions of parameters could not have been trained without the usage of SGD-based optimizers. 

Despite its versatility and excellent performance, SGD demands a substantial amount of iterations to converge when solving complex problems over a large number of parameters. Consequently, training a large model with SGD may be a time-consuming task \cite{cheng2018model}. To mitigate this issue, it is common to distribute the computations performed by the SGD optimizer across multiple CPUs, GPUs, or even across multiple compute nodes. Various settings used to distribute the SGD optimizations are described in the literature~\cite{ouyang2020communication, mayer2020scalable}; in this paper, we will focus on {\em data parallelism}, which is the most common form of distributed training for SGD-based optimizers \cite{mayer2020scalable}. 

When performing data-parallel training, multiple workers are initialized with the same parameters (i.e., weights), and they simultaneously calculate their models' gradients based on different data samples obtained from the training data \cite{ouyang2020communication}. In a {\em distributed} (but synchronous) {\em SGD} setting, the gradients of all workers are aggregated periodically and used to update a centralized model stored in a single parameter server which is then broadcast to all workers \cite{mayer2020scalable}. This setting however has two main disadvantages: (1) synchronization idle times, which occur whenever a worker finishes calculating its gradient faster than other workers and it spends resources waiting for delayed workers \cite{mayer2020scalable}; and (2) bottlenecks in the parameter server, which may occur due to all workers communicating with the same centralized server simultaneously \cite{chen2019round}. To further reduce the training time when using SGD-based optimizers, researchers therefore actively studied how to overcome these issues coming from synchronization and centralization.

To avoid synchronization barriers among workers, current efforts are mostly focused on {\em asynchronous SGD} (ASGD). In an ASGD setting, workers do not wait for updates from the parameter server to resume computing new gradients \cite{ouyang2020communication}. Therefore, they eliminate the workers' synchronization barriers and, consequently, idle times. However, as a consequence, the parameter server may receive also delayed gradients calculated during past iterations, which complicates the ASGD convergence proof. Very recently, Koloskova, Stich, and Jaggi \cite{koloskova2022sharper} proved an ASGD convergence rate of $\mathcal{O}(\frac{\sigma^2}{\epsilon^2}) + \mathcal{O}(\frac{\sqrt{\tau_{avg}\tau_{max}}}{\epsilon})$ to an $\epsilon$-small error with $\tau$ representing the gradient delay; and $\mathcal{O}(\frac{\sigma^2}{\epsilon^2}) + \mathcal{O}(\frac{\tau_{avg}G}{\epsilon^{3/2}}) + \mathcal{O}(\frac{\tau_{avg}}{\epsilon})$ when the norm of the gradients is additionally also bounded by $G$. These results are very broadly applicable because, based on them, the authors could also prove that ASGD always converges faster than mini-batch SGD.

To avoid these bottlenecks arising from centralized parameter servers, one may choose a decentralized setting, in which it is not necessary to aggregate gradients in a centralized parameter server. Instead, workers exchange gradients directly among themselves, aggregate them locally, and update themselves accordingly \cite{mayer2020scalable}. The exchange of gradients among workers may be performed based on different network topologies. Common such topologies are the {\em fully connected topology}, in which all workers are pairwisely connected; and the {\em ring topology}, in which the workers are connected in the form of a closed loop or ring, with each worker communicating to two adjacent workers \cite{mayer2020scalable}.

Algorithms that consider ASGD in a decentralized topology are still rare due to the complexity of proving its convergence under both of these relaxations simultaneously. Most authors have studied the convergence rate of decentralized ASGD for convex functions \cite{wu2017decentralized, assran2020asynchronous, jiang2021asynchronous} and a few others proved convergence rates also for non-convex ones \cite{lian2018asynchronous}. Nevertheless, most of those proofs still rely on partial synchronization during models communication and specific network topologies.

\subsection{Contributions}
\squishlist
\item We formally prove a convergence rate for DASGD of $\mathcal{O}(\frac{\sigma}{\epsilon^2}) + \mathcal{O}(\frac{QS_{avg}}{\epsilon^{3/2}}) + \mathcal{O}(\frac{S_{avg}}{\epsilon})$ to an $\epsilon$-small error, with $Q$ bounding the norm of the gradients, and $S_{avg}$ representing the {\em average staleness} among all models. Here, staleness is defined as the pairwise symmetric difference between the sets of gradients calculated by one model and applied by another model (cf. Section \ref{sec:convergence-analysis}). This convergence rate is guaranteed for any non-convex, L-smooth, and homogeneous objective function with bounded gradients and for a constant stepsize of $\eta \leq (4LS_{avg})^{-1}$.
\item With no assumption over the size of the gradients, we additionally prove a convergence rate for DASGD of $\mathcal{O}(\frac{\sigma}{\epsilon^2}) + \mathcal{O}(\frac{\sqrt{\hat{S}_{avg}\hat{S}_{max}}}{\epsilon})$ to an $\epsilon$-small error, using a constant stepsize of $\eta \leq (4L\sqrt{\hat{S}_{avg}\hat{S}_{max}})^{-1}$. Here, $\hat{S}_{avg}$ and $\hat{S}_{max}$ represent a loose version of the {\em average} and {\em maximum staleness} among the distributed models, respectively.
\item We empirically demonstrate how staleness impacts the convergence rate of a logistic regression, a quadratic function, and a Convolutional Neural Network (CNN) model when optimized with DASGD. 
\squishend

\subsection{Limitations}
\squishlist
\item Our proof currently does not contemplate varying stepsizes. 
\item The stepsize shall be defined based on the staleness, which may not be known before the training. 
\item Our proof does not currently contemplate heterogeneous functions, commonly observed in federated learning problems. Nevertheless, the effect of the heterogeneity should be orthogonal to the effect of our staleness measure.
\squishend

\section{Related Works}

\smallskip\noindent\textbf{Decentralized SGD.~} The idea that decentralized SGD (DSGD) can outperform centralized SGD has been shown in \cite{lian2017can}. However, depending on the topology of the underlying network, the communication cost when not relying on a centralized parameter server can increase quadratically with the number of distributed worker nodes $n$ \cite{mayer2020scalable}. One approach to lower the communication cost is to use so-called ``gossip algorithms'' \cite{lian2017can, koloskova2019decentralized, koloskova2020unified}, which allow workers to exchange and aggregate their gradients only with their immediate neighbors (usually 2 to 3), instead of all worker nodes. Then, those neighbors disseminate the messages received across the network iteratively, and after approximately $1.639\,\log_2(n)$ communication steps all workers will receive the initial message \cite{ben2019demystifying}. However, networks with more than 32 workers may suffer in terms of convergence and performance due to this relatively high delay in the exchange of messages \cite{ben2019demystifying}.
 
\smallskip\noindent\textbf{Asynchronous SGD.~} Introduced already decades ago, the study of asynchronous SGD (ASGD) has gained an increasing amount of attention again recently. At first, works such as Hogwild! \cite{recht2011hogwild} focused on proving the convergence of ASGD under the assumption of the sparseness of the optimized models. Recently, less restricted convergence proves were developed, which also contemplate dense models as \cite{koloskova2022sharper, mishchenko2022asynchronous, cohen2021asynchronous, arjevani2020tight, xie2019asynchronous, nguyen2022federated, toghani2022unbounded, stich2020error, stich2018local}. A main aspect to facilitate the proof of convergence used by \cite{mishchenko2022asynchronous} is to scale the gradients based on their delay, such that delayed gradients have less impact when updating the model. Another technique seen in \cite{cohen2021asynchronous} is to simply discard gradients which are too delayed. Both techniques described above are easy to implement in a centralized setting, in which a parameter server can adapt its learning rate based on the delay of the gradient that it applies. However, this adaptation is not straightforward in a decentralized setting, in which there is no server to decide how delayed the gradients are. Furthermore, part of these ASGD convergence proofs do not cover the optimization of non-convex functions \cite{arjevani2020tight, xie2019asynchronous, stich2018local}. In addition, some proofs also rely on varying stepsizes, which may hinder convergence proofs under a decentralized setting (cf. Section \ref{sec:convergence-analysis}). Thus, only very few convergence proofs provided in the literature \cite{koloskova2022sharper,nguyen2022federated,toghani2022unbounded, stich2020error} cover ASGD with fixed learning rates and non-convex objective functions. We highlight here the recent proof by Koloskova, Stich and Jaggi \cite{koloskova2022sharper} which provides the currently best convergence rate for ASGD under mild assumptions, while considering a fixed learning rate and non-convex functions. Specifically, they were able to prove that ASGD convergence does not depend on the maximum delay of gradients when those gradients are bounded. 

\smallskip\noindent\textbf{Decentralized and Asynchronous SGD.~} To mitigate the bottleneck from centralized parameter servers and avoid idle time from synchronizing distributed models, some authors focused their research on decentralized and asynchronous SGD \cite{wu2023delay,ram2010asynchronous, sirb2016consensus,assran2020asynchronous,wu2017decentralized,even2021asynchronous,srivastava2011distributed,ram2009asynchronous,lian2018asynchronous}. However, most of those studies make strong assumptions as \cite{wu2023delay,ram2010asynchronous, sirb2016consensus,wu2017decentralized,assran2020asynchronous,even2021asynchronous,srivastava2011distributed}, which do not prove convergence for non-convex functions, or they prove the convergence of their models only when the number of iterations goes to infinity \cite{wu2023delay,srivastava2011distributed, ram2009asynchronous}. Contrarily, \cite{lian2018asynchronous} proved the convergence rate of decentralized and asynchronous SGD for non-convex functions by relying on specific network topologies and atomic communication among nodes, which requires a partial synchronization between any two nodes during the optimization process, hindering the convergence time in practice \cite{jiang2021asynchronous, luo2020prague, miao2021heterogeneity}. In a concurrent work, \cite{DBLP:conf/iclr/BornsteinRWBH23} reached a similar convergence rate to ours by also using a wait-free approach that eliminates the partial synchronization and special network architecture requirements. Their optimization process relies on a novel client-communication matrix that governs the models averaging. Differently, we freely exchange gradients among models and introduce the new {\em staleness} metric to measure the difference between those models. At last, both approaches were able to eliminate from the convergence rate the dependency from the slowest model in the network. We eliminate this dependency by assuming gradients are bounded while they eliminate it by defining a minimum amount of iterations until convergence and choosing their learning rate based on it.

A detailed comparison of all papers discussed above under these aspects can be found in Table \ref{tab:related_works}.

\begin{table}[]
\small
    \centering
    \begin{tabular}{c|cccc}
        \textbf{References} & \textbf{Asynchronous} & \textbf{Decentralized} & \multicolumn{1}{c}{\begin{tabular}[c]{@{}c@{}}\textbf{Non-convex}\\ \textbf{functions}\end{tabular}} &  \multicolumn{1}{c}{\begin{tabular}[c]{@{}c@{}}\textbf{Convergence}\\ \textbf{rate}\end{tabular}} \\ \hline\rule{0pt}{1\normalbaselineskip}
        \cite{lian2017can, koloskova2020unified,koloskova2019decentralized, cohen2021asynchronous} & No & Yes & Yes& Yes \\
        \cite{koloskova2019decentralized} & No & Yes & No &Yes \\
        \cite{koloskova2022sharper, nguyen2022federated, stich2020error, mishchenko2022asynchronous, cohen2021asynchronous, toghani2022unbounded} & Yes & No & Yes &Yes \\
        \cite{recht2011hogwild, xie2019asynchronous, stich2018local, arjevani2020tight} & Yes & No & No & Yes \\
        \cite{wu2023delay, srivastava2011distributed} & Yes & Yes & No & Infinity \\
        \cite{ram2010asynchronous, sirb2016consensus, wu2017decentralized,assran2020asynchronous,even2021asynchronous} & Yes & Yes & No & Yes \\
        \cite{ram2009asynchronous} & Yes & Yes & Yes & Infinity \\
        \textbf{Theorem \ref{theorem}, \cite{lian2018asynchronous}, \cite{DBLP:conf/iclr/BornsteinRWBH23}} & \textbf{Yes} & \textbf{Yes} & \textbf{Yes} & \textbf{Yes}
        
    \end{tabular}
    \caption{Comparison between SGD convergence proofs.}
    \label{tab:related_works}
\end{table}

\section{Optimization Objective \& DASGD Algorithm}
In this section, we present the setup under which we prove an upper bound of the convergence rate of {\em decentralized and asynchronous SGD} (DASGD). First, we formally describe the optimization problem solved and the assumptions made. Then, we present the characteristics expected from the network topology and communication strategy on which our convergence analysis is based. Finally, we present the algorithm describing the protocol followed by each of the decentralized worker nodes. 

Our setting is based on the one presented in \cite{koloskova2022sharper}, which provides tight convergence rates for ASGD under various settings. We in particular generalize their analysis of ASGD under a homogeneous setting with a fixed stepsize (their Theorem 6), and extend this to a decentralized setting which no longer relies on a centralized parameter server. The main intuition behind our approach is that, if we can assume that any gradient computed by any of the local models also eventually reaches all the other models, then each of the local models effectively works like a parameters server itself, and therefore all local models will converge to the same global model. This allows us to develop analogous constructions to the proofs provided in \cite{koloskova2022sharper} and \cite{nguyen2022federated} for our decentralized setting.

\subsection{Optimization Objective}
Following \cite{koloskova2022sharper}, we consider the {\em optimization objective} shown below:
\begin{equation}
    \min_{x \in \mathbb{R}^d} \left[ f(x) \coloneqq \frac{1}{n} \sum^{n}_{i=1} \left[ f_i(x) = \mathbb{E}_{\xi \sim \mathcal{D}} \, F_i(x, \xi) \right] \right]
\end{equation}
Here, $F_i$ represents a local loss function which is accessed by node $i$ with parameters (i.e., weights) $x$ on data samples $\xi \sim \mathcal{D}$, considering $i \in \{1,\ldots,n\}$ (in the following abbreviated as $i \in [n]$). 
As in \cite{koloskova2022sharper}, each $f_i(x)$ with $f_i : \mathbb{R}^d \rightarrow \mathbb{R}$ is assumed to be a stochastic function, i.e., $f_i(x) = \mathbb{E}_{\xi \sim \mathcal{D}_i} \, F_i(x, \xi)$, and is accessed only via its local gradients $\nabla F_i(x, \xi)$. 

We remark that this is a very generic setting, which captures any data distribution $\mathcal{D}$ and applies to any (smooth, but possibly non-convex) objective function. In case the optimization problem, for example, is deterministic, we may set $f_i(x) = F_i(x, \xi), \forall{\xi}$; if, on the other hand, $\mathcal{D}$ is a uniform distribution with local samples $\{\xi_i^1,\ldots, \xi_i^{m_i}\}$, we have $f_i(x) = \frac{1}{m_i} \sum^{m_i}_{t=1} F_i(x, \xi_i^t)$. However, as opposed to \cite{koloskova2022sharper}, we assume all samples $\xi$ to come from the sample global distribution $\mathcal{D}$ rather than allowing different local distributions $\mathcal{D}_i$ (which, together with fixed stepsizes, matches the setting of Theorem 6 in \cite{koloskova2022sharper}). Consequently, we also only look at homogeneous functions $f_i(x)$, as stated in Assumption \ref{as:func-heter} below.

\subsubsection{Notation}
\label{subsec:Notation}
Below we summarize the notation we use to describe our setup (and also later refer to in Appendix \ref{sec:Appendix} for our proof of Theorem \ref{theorem}):
{
\squishlist
\item $\eta$, fixed learning rate;
\item $n$, number of distributed worker nodes and models;
\item $x^0$, initial model (distributed across all worker nodes);
\item $x^t_i$, local model $i$ at iteration $t$, with $x^0_i = x^0$ and $i \in [n]$;
\item $x^t$, any model $x^t_i, \forall i \in [n]$;
\item $g^t_i$, gradient calculated using model $x^t_i$ and sample $\xi^t_i$; thus, $g^t_i = \nabla F(x^t_i, \xi^t_i)$;
\item $G^t_i$, set of gradients applied from $x^0$ to $x^t_i$; thus, $x^t_i = x^0_i - \eta \sum_{g \in G^t_i}g$;
\item $S_{i,j}^{t,s}$, staleness between set of gradients $G^t_i$ and $G^s_j$ (cf. Definition \ref{def-staleness});
\item $\hat{S}_{i,j}^{t,s}$, loose version of staleness between $G^t_i$ and $G^s_j$ (cf. Definition \ref{def-staleness});
\item $\delta_{i,j}^{t,s}$, sum of gradients in $S_{i,j}^{t,s}$ weighted by $\eta$; thus, $\delta_{i,j}^{t,s} = \eta \sum_{g \in S_{i,j}^{t,s}} g$, with $\delta_{i,j}^{t,s} \in \mathbb{R}^d$;
\item $\Delta_{i,j}^{t,s}$, upper bound of $\delta_{i,j}^{t,s}$; $\Delta_{i,j}^{t,s} = \eta\sum_{g \in S_{i,j}^{t,s}} |g|$, with $\Delta_{i,j}^{t,s} \in \mathbb{R}^d$.
\squishend
}
Throughout this paper, we refer to L$_2$-norm $\|\cdot\|_2$ as our default norm for vectors and thus simplify our notation by writing $\|\cdot\|$. Moreover, with $|v|$, for a vector $v \in \mathbb{R}^d$, we denote a corresponding vector consisting of the absolute values along $v$'s dimensions, i.e., $|v| = \langle |v_1|,|v_2|, \ldots, |v_d|\rangle$.

\subsubsection{Assumptions}
\label{subsec:Assumptions}
The following assumptions are considered throughout the description of our setting and the convergence analysis provided in Appendix \ref{sec:Appendix}.

\smallskip
\begin{assumption}[Bounded variance]\label{as:bounded-variance} \emph{There exists a constant $\sigma$, such that:}
\begin{equation}
    \mathbb{E}_{\xi \sim \mathcal{D}} \, \| \nabla F_i(x, \xi) - \nabla f_i(x) \|^2 \leq \sigma^2 \hspace*{40pt}  \forall i \in [n], \forall x \in \mathbb{R}^d
    \label{eq:bounded-variance}
\end{equation}
\end{assumption}

\smallskip
\begin{assumption}[Function homogeneity]\label{as:func-heter} \emph{The functions $f_i$ are homogeneous, thus:}
\begin{equation}
    \|\nabla f_i(x) - \nabla f_j(x)\| = 0 \hspace*{40pt} \forall i,j \in [n], \forall x \in \mathbb{R}^d
    \label{eq:func-heter}
\end{equation}
\end{assumption}

\smallskip
\begin{assumption}[Lipschitz gradient]\label{as:l-smooth} \emph{The gradient is L-smooth and there exists a constant $L \geq 1$, such that:}
\begin{equation}
    \| \nabla f_i(y) - \nabla f_i(x)  \| \leq L \|x-y\| \hspace*{40pt}  \forall i \in [n], \forall x,y \in \mathbb{R}^d
    \label{eq:l-smooth}
\end{equation}
\end{assumption}

\smallskip
\begin{assumption}[Bounded gradient]\label{as:bounded-gradient} \emph{There exists a constant $Q \geq 0$, such that:}
\begin{equation}
    \| \nabla f_i(x) \|^2 \leq Q^2 \hspace*{40pt}  \forall i \in [n], \forall x \in \mathbb{R}^d
    \label{eq:bounded-gradient}
\end{equation}
\end{assumption}

\smallskip
The above assumptions are very common in the context of SGD and have been adopted by various classical works. Specifically, Assumptions \ref{as:bounded-variance} and \ref{as:l-smooth} are commonly used for most SGD proofs. The only restriction we make when compared to other papers is to consider $L \geq 1$ (instead of the commonly used $L \geq 0$), which allows us to perform simplifications during our proof that guarantee the provided convergence rate without further theoretical or practical implications. For Assumption \ref{as:bounded-gradient}, we adopt the same strategy as \cite{koloskova2022sharper} by providing two different bounds, one considering this assumption and one not considering it. At last, considering Assumption \ref{as:func-heter} (also explored by \cite{arjevani2020tight, stich2020error, agarwal2011distributed,feyzmahdavian2016asynchronous, lian2015asynchronous, sra2016adadelay}), we focus our analysis on distributed gradient computations as they are typically performed in a data center \cite{dean2012large} and in which all worker nodes (using CPUs or GPUs) process data from the same distribution $\mathcal{D}$, and, consequently, guarantee Assumption \ref{as:func-heter}.

\subsection{Network Topology \& Communication Protocol}

Our convergence proof for DASGD is flexible enough to allow any network topology or communication protocol between the worker nodes (e.g., fully connected, ring, and mesh topologies {\cite{mayer2020scalable}) as long as the following characteristics are respected. 
\squishlist
\item {\em connected graph}: there must exist a communication path between any two nodes in the network;
\item {\em non-lost messages}: a message (encoding gradients) that is sent by a worker node shall eventually be received by all other nodes in the network;
\item {\em non-repeated messages}: each message will be received at most once by each worker node.
\squishend

We remark that neither our DASGD algorithm (cf. Section \ref{sec:DASGDalg}) nor our convergence analysis (cf. Section \ref{sec:convergence-analysis}) depend on the order of the sent messages to be preserved among the worker nodes. We see this as a strong feature of our approach which enhances the flexibility of the communication protocol that may be adopted. In practice, the longer it takes for two nodes to exchange their local gradients, the larger the staleness among their models will become and the worse the bounds for the convergence rate will be (cf. Section \ref{sec:convergence-analysis}). Moreover, as our messages are timestamped (using local step counters $t$ only), our staleness measure immediately complies also with unordered messages, since we can exactly determine the amount of lagging messages from these step counters. However, in practice, one will observe better convergence rates when using denser network topologies with a frequent and ordered communication between the worker nodes.

\subsection{Decentralized \& Asynchronous SGD Algorithm}
\label{sec:DASGDalg}
We next introduce our {\em DASGD algorithm}, as summarized in Algorithm \ref{alg:model}. From a practical perspective, the idea behind it is to calculate new gradients only if no updates (i.e., gradients received from other models) are available. In doing so, we reduce the dissimilarity between the asynchronous models throughout the training. Specifically, we eliminate idle times coming from synchronization barriers between nodes (common on synchronous SGD) and prevent bottlenecks on parameter servers (common on centralized SGD). 

\begin{algorithm}[ht]
\small
\SetAlgoLined
\DontPrintSemicolon
\SetKwFunction{FMain}{initializeModel}
    \SetKwProg{Fn}{Constructor}{:}{}
    \nl \Fn{\FMain{$x^0$, $i$}}{
        \nl $t = 0$;\;       
        \nl $x^t_i = x^0$;\;
    }
    
    \SetKwProg{Fn}{Function}{:}{}
    \nl	\Fn{{\tt trainModel}}{
    \nl \While{true}{
            \nl $g = \texttt{ receiveGradient}()$;\quad\CommentSty{//returns NULL if there is no incoming gradient}\; 
            \nl \eIf{$g \neq \text{NULL}$}{ 
                \nl $x^{t+1}_i = x^{t}_i - \eta g$;\;
                \nl $t = t + 1$;\;
            }{
                \nl \If{$\; \exists \; \xi^t_i$}{ \quad\CommentSty{//only then compute a new gradient}\;
                    \nl $g^t_i = \nabla F(x^t_i, \xi^t_i)$;\;
                    \nl $x^{t+1}_i = x^t_i - \eta g^t_i$;\;
                    \nl $\texttt{sendGradient}(g^t_i)$; \quad\CommentSty{//non-blocking operation}\;
                    \nl $t = t + 1$;\;
                }
            }
    }    
    }
 \caption{DASGD Algortihm.}
 \label{alg:model}
\end{algorithm}

In Algorithm \ref{alg:model}, we show the protocol performed by all $i \in [n]$ worker nodes. First, all nodes are initialized with the same model parameters $x^0$, their node id $i$, and a local iteration counter $t = 0$. 

During training, on Lines 6 and 7, node $i$ checks if any gradient $g$ computed by another node is available. If so, it immediately (i.e., asynchronously) updates itself based on the received gradient and by using a fixed learning rate $\eta$ (Line 8). It then also increments its iteration counter $t$ by 1 (Line 9) and resumes its training back on Line 5. 

If, on the other hand, no gradients were received, worker $i$ checks if a new training sample $\xi^t_i$ is available (Line 10). It then calculates a new gradient $g^t_i$ based on its current model's weights $x^t_i$ and iteration counter $t$ (Line 11). After calculating $g^t_i$, worker $i$ updates itself based on this gradient and the learning rate $\eta$ (Line 12). It then sends (i.e., in the simplest case ``broadcasts'') this gradient to the other workers it is connected to in the network (line 13). At last, it increments its iteration counter $t$ by 1 (Line 14); and it resumes its training back on Line 5. 

The training is performed until $\nexists \, \xi^t_i, \forall i \in [n]$ and all gradients $g^t_i$ were applied by all nodes $i, \forall i \in [n]$, which assumes only a single (and final) synchronization point among all worker nodes.

\section{Convergence Rate Analysis}
\label{sec:convergence-analysis}

Throughout the execution of Algorithm \ref{alg:model}, a local model $x_i$ will update itself using gradients calculated by a model $x_j$, with $i, j \in [n]$. As opposed to current works, considering our asynchronous and decentralized setting, we cannot guarantee that $x_i$ and $x_j$ represent the same model at different iteration points. Thus, neither
$x_i^t = x_j^{t-\tau}$ nor $x_j^t = x_i^{t-\tau}, \forall \tau \in \mathbb{N}$ are assured, considering $t$ as the model's iteration and $\tau$ a possible delay. Strictly speaking, we cannot even assume that the iteration counters $t$ at each model are synchronized. Therefore, we rely on a different approach to calculate the difference between models. We start by representing a model $x^t_i$ via a set of gradients $G^t_i$ it either received or computed itself at iteration $t$.

\begin{remark} The set of gradients $G^t_i$ uniquely represents the evolution of model $x^0$ into $x^t_i$ when using a fixed learning rate $\eta$. Thus, we can describe $x^t_i$ as:
\begin{equation}
x^t_i = x^0 - \eta \sum_{g \in G^t_i} g
\end{equation}
\end{remark}

Therefore, we can quantify the dissimilarity between models based on the set of gradients which they applied to themselves since they were initialized, provided that these were initialized with the same parameters $x^0$ and updated with the same learning rate $\eta$. More specifically, we calculate the symmetric difference between these two sets, which produces the {\em staleness} of gradients \cite{tosi2022convergence}, represented here as $S^{t,s}_{i,j}$, between pairs of models $i$, $j$ and iterations $t$, $s$ (cf. Figure \ref{fig:staleness}). Our convergence bounds depend on the average and the maximum sizes of these staleness sets, represented as $S_{avg}$ and $S_{max}$, respectively.

\begin{figure}[h]
\centering
\includegraphics[width=0.7\textwidth]{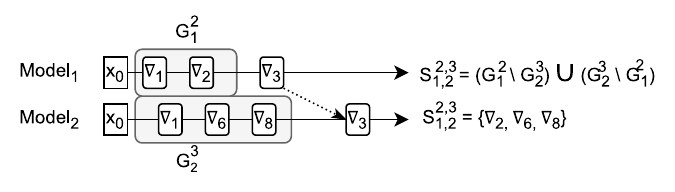}
\caption{{\em Staleness} $S^{2,3}_{1,2}$ at the time when Model$_2$ applies gradient $\nabla_3$ calculated by Model$_1$. The grey areas determine the sets of gradients $G_1^2$ and $G_2^3$ used to calculate the symmetric difference. }
\label{fig:staleness}
\end{figure}

For the following definition, let $x_j^s$ represent model $j$ that calculated a gradient $g_j^s$ at iteration $s$, and let $x_i^t$ represent the model $i$ that applies $g_j^s$ at iteration $t+1$. We then represent $x_j^s$ and $x_i^t$ using $G^s_j$ and $G^t_i$ respectively, assuming that both models were initialized with the same parameters $x^0$ and updated with the same learning rate $\eta$. 

\begin{definition}[Staleness]
\label{def-staleness}
Under Assumption \ref{as:bounded-gradient} (bounded gradients), we define {\em staleness} $S^{t,s}_{i,j}$ as the symmetric difference between the sets of gradients $G^t_i$ and $G^s_j$ applied by models $i$ and $j$ at steps $t$ and $s$, respectively:
\begin{equation}
S^{t,s}_{i,j} = (G^t_i \setminus G^s_j) \cup (G^s_j \setminus G^t_i) \hspace*{60pt} (\text{when Assumption \ref{as:bounded-gradient} holds})
\end{equation}

Moreover, for our convergence proof in the case when Assumption \ref{as:bounded-gradient} (bounded gradients) does not hold, we also define a looser version of staleness $\hat{S}^{t,s}_{i,j}$. This version is defined as the union of (1) the symmetric difference between the sets of gradients $G^t_i$ and $G^s_j$ (as before); and (2) the staleness $\hat{S}^{t,u}_{i,k}$, which recursively also includes gradients $G^u_k$ calculated by another model $k$ at step $u$ (and which belong to $G^s_j$ but not to $G^t_i$):
\begin{equation}
\hat{S}^{t,s}_{i,j} = (G^t_i \setminus G^s_j) \cup (G^s_j \setminus G^t_i) \cup_{g^u_k \in (G^s_j \setminus G^t_i)} \hat{S}^{t,u}_{i,k} \hspace*{30pt} (\text{when Assumption \ref{as:bounded-gradient} does not hold})
\end{equation}
\end{definition}

We highlight here that, when calculating the staleness between models $x^t_i$ and $x^s_j$ (using $G^t_i$ and $G^s_j$), each model requires only a local iteration counter, i.e., $0 \leq t \leq T_i$ and $0 \leq s \leq T_j$ (cf. Algortihm \ref{alg:model}). To simplify further notations, we thus refer to $T$ as the maximum step counter $T_i$ of the model that is currently under consideration. When analyzing the staleness $S^{t,s}_{i,j}$ between pairs of models $i$, $j$, then by convention $T$ will represent the step counter of the left-hand model (in this case, and throughout the rest of the paper, the one of model $i$).


\begin{definition}[Average \& Maximum Staleness]
Let $g^s_j$ be the gradient applied to model $i$ at step $t+1$. Then, $|S^{t,s}_{i,j}|$ and $|\hat{S}^{t,s}_{i,j}|$ denote the sizes of the {\em staleness} sets between model $i$ at step $t$ and model $j$ at step $s$. Consequently, we define the {\em maximum} and {\em average staleness} among all models in the same manner for $S$ and $\hat{S}$: 
\begin{equation}
S_{max} := \max_{0 < i \leq n,\,0 \leq t \leq T}\{|S^{t,s}_{i,j}|\} \hspace*{15pt} \text{and} \hspace*{15pt} S_{avg} := \max_{0 < i \leq n}\{\frac{1}{{(T+1)}}\sum_{t = 0}^T |S^{t,s}_{i,j}|\}
\label{eq:s_avg}
\end{equation}

\begin{equation}
\hat{S}_{max} := \max_{0 < i \leq n,\,0 \leq t \leq T}\{|\hat{S}^{t,s}_{i,j}|\} \hspace*{15pt} \text{and} \hspace*{15pt} \hat{S}_{avg} := \max_{0 < i \leq n}\{\frac{1}{{(T+1)}}\sum_{t = 0}^T |\hat{S}^{t,s}_{i,j}|\}
\label{eq:s_avg*}
\end{equation}

\label{def:s_avg}
\end{definition}


The key idea behind our DASGD approach comes from the observation that gradient applications with a fixed learning rate are both {\em associative} and {\em commutative}. 
Therefore, we can guarantee convergence among multiple decentralized models as long as (1) the models are initialized with the same weights; (2) the models apply the same gradients (independently of their order); (2) and the same gradients are applied with the same learning rate. Moreover, by considering our definition of staleness, we can estimate the dissimilarity between models and thereby determine the expected convergence rate of the global model.

Below, we present the central results of our convergence analysis.
\begin{theorem}
\label{theorem}
Considering Assumptions \ref{as:bounded-variance}, \ref{as:func-heter}, \ref{as:l-smooth}, \ref{as:bounded-gradient}, and a constant stepsize $\eta \leq \frac{1}{4LS_{avg}}$, Algorithm \ref{alg:model} reaches $\frac{1}{T+1} \sum^{T}_{t=0}( \| \nabla f(x^{t})\|^2) \leq \epsilon$ after 
\begin{equation}\label{eq:1theorem}
\mathcal{O}(\frac{\sigma}{\epsilon^2}) + \mathcal{O}(\frac{QS_{avg}}{\epsilon^\frac{3}{2}}) + \mathcal{O}(\frac{S_{avg}}{\epsilon}) \hspace*{40pt} \text{iterations}.
\end{equation}

Moreover, without Assumption \ref{as:bounded-gradient} and using $\eta \leq \frac{1}{4L\sqrt{\hat{S}_{avg}\hat{S}_{max}}}$, Algorithm \ref{alg:model} reaches $\frac{1}{T+1} \sum^{T}_{t=0}(\|\nabla f(x^{t})\|^2) \leq \epsilon$ after
\begin{equation}\label{eq:2theorem}
\mathcal{O}(\frac{\sigma^2}{\epsilon^2}) + \mathcal{O}(\frac{\sqrt{\hat{S}_{avg}\hat{S}_{max}}}{\epsilon}) \hspace*{60pt} \text{iterations}.
\end{equation}
\end{theorem}

Our detailed proof of Theorem \ref{theorem} is available in Appendix \ref{proof}.

\subsection{Discussion}


\smallskip\noindent\textbf{DASGD vs. SGD.~} As seen in Theorem \ref{theorem}, the importance of the staleness decreases over time. Therefore, we can conclude that if $S_{max} \leq \sqrt{T}$ or $\hat{S}_{max} \leq \sqrt{T}$, DASGD converges at a similar rate as synchronous SGD. Nevertheless, DASGD eliminates idle time coming from slower nodes, which makes it calculate gradients at a higher pace and, consequently, converging faster than synchronous approaches. 

\smallskip\noindent\textbf{Network topology.~} The convergence of DASGD is directly impacted by $S_{avg}$ or by $\hat{S}_{avg}$ and $\hat{S}_{max}$ (when Assumption \ref{as:bounded-gradient} does not hold), which depend on three factors: (1) network topology; (2) communication latency; and (3) computational resources available among nodes. Assuming a scenario with computational resources being equally distributed and the latency being smaller than the computation time it takes to calculate gradients, we can directly measure the impact of the network topology chosen. For example, in the above mentioned conditions, in a fully connected topology with $n$ nodes, one can expect $S_{avg} = \frac{n+1}{2}$ and $S_{max} = n$. On the other hand, in a ring topology, messages will take longer to reach their destinations (up to $n$ times longer), with $S_{avg} = \frac{n^2+1}{2}$ and $S_{max} = n^2$. 

\smallskip\noindent\textbf{Tightness.~} When compared to a centralized setting, the sizes of our {\em staleness} sets behave analogously to the {\em delay} used in works such as \cite{koloskova2022sharper,nguyen2022federated}. Therefore, we can also interpret the delay as a special case of our staleness, the latter representing the size of the symmetric difference of gradient sets obtained from the same model (i.e., the one at the parameter server) across different iterations. Considering this, we can view the convergence proof provided in this paper as a generalization of Theorem 6 presented in \cite{koloskova2022sharper}, which in turn also extends traditional mini-batch SGD with constant stepsizes. This indicates the tightness of the given convergence rate, coinciding with the known lower bound for mini-batch SGD of $\Theta(\frac{\sigma^2}{n\epsilon^2} + \frac{1}{\epsilon})$ batches which are necessary to reach $\epsilon$ as stationary point (when setting $S_{max} = n$ and $S_{avg} = \frac{n}{2}$ in Equation (\ref{eq:2theorem})).

\smallskip\noindent\textbf{Practical limitations.~} Despite the simplicity of Algorithm \ref{alg:model}, there still are practical limitations when choosing the learning rate $\eta$. First, $\eta$ shall be fixed throughout the training, which is known not to be optimal in practice. Second, $\eta$ shall be bounded by $S_{avg}$ or by $\hat{S}_{max}$ and $\hat{S}_{avg}$, which may not be known beforehand. Nevertheless, in practice, one can estimate those values and reach convergence under the provided bounds without further concerns, as seen in our experiments in Section \ref{sec:experiments}.

\section{Experiments}
\label{sec:experiments}

In this section, we show how the theoretical bounds for DASGD introduced in Theorem \ref{theorem} traverse to practical experiments. 
The experiments were run on a DELL PowerEdge R840 server with 192 cores using an MPI environment which simulates multiple ranks with no shared memory. 

Inspired by \cite{koloskova2022sharper}, we assessed DASGD by performing our first two experiments in a scenario with no stochastic noise, i.e., $\sigma = 0$. In doing so, the convergence rate of the model being optimized is reduced to $\mathcal{O}(\sqrt{\hat{S}_{avg}\hat{S}_{max}}~\epsilon^{-1})$, thus depending only on the {\em staleness} observed during the optimization. In both experiments, we fixed $n=2$, $\epsilon=1e^{-12}$, $\eta=0.002$ and varied $\hat{S}_{max}$ from $0$ to $100$, which in particular maintains $\eta < (4L\sqrt{\hat{S}_{avg}\hat{S}_{max}})^{-1}$. 

To artificially control the staleness during the optimization, we followed two strategies: (1) we reduced the gradient calculation speed of one of the models, making it $x$ times slower; (2) we artificially increased the time it took to calculate each gradient by $0.05$ seconds. The first technique guarantees that one of the workers outperforms the other one, thereby increasing the {\em maximum staleness}. The second technique guarantees that the gradient calculation takes orders of times more than the actual gradient application, which gives the slower model the chance to calculate new gradients instead of being overloaded by the application of the gradients calculated by the fastest model (which is the case in most real-world use-cases). Thus, with both techniques described above and the other parameters fixed, we can expect $\hat{S}_{max} = x$ during the optimization. Furthermore, we estimated the error of the optimized functions based on the average of the L$_2$-norm of their gradients over the last 30 iterations. We chose to optimize the same functions as in \cite{koloskova2022sharper}:
\squishlist
\item a {\em quadratic function} $f(x) = \frac{1}{2} \|Ax-b\|^2$, with $x, b \in \mathbb{R}^{10}$, $i \in [1,10]$, $b_i \sim \mathcal{N}(0,1)$, and $A \in \mathbb{R}^{10x10}$ being a random matrix with $\lambda_{min}(A) = 1$, $\lambda_{max}(A) = 2$ (cf. Figure \ref{fig:logistic_regression});
\item a {\em logistic-regression function} $f(x) = \frac{1}{m}\sum^m_{j=1}\log(1+ \exp(-b_ja^T_jx))$, with $a_j \sim \mathcal{N}(0,1)^{20}$, $x \in \mathbb{R}^{20}$, $m = 100$, and $b_j$ is sampled uniformly at random from $\{-1,1\}$ (cf. Figure \ref{fig:quadratic_function}).
\squishend

In addition, we analyzed the impact of staleness on the convergence rate and time of a Convolutional Neural Network (CNN) for image classification \cite{cnn}, which we trained using the CIFAR-10 dataset \cite{krizhevsky2009learning}. In this experiment, we set $\epsilon = 1$, $\eta = 1e^{-4}$, and varied $n$ between $1$ and $25$. By varying $n$, we ended up increasing $\hat{S}_{max}$ indirectly---close to what we may expect in a real-world scenario. The results of this experiment can be seen in Figure \ref{fig:cifar}.

\begin{figure}[h]
\center
  \hspace{-6mm}
  \begin{subfigure}[b]{0.33\textwidth}
    \centering
    \includegraphics[width=\textwidth]{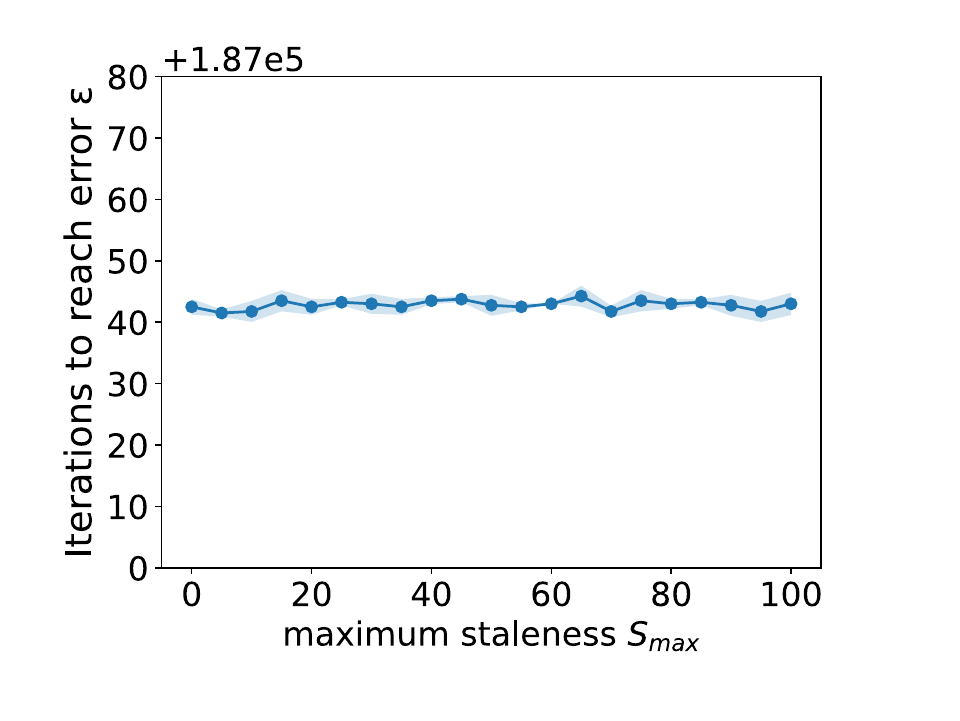}
    \caption{Logistic regression}
    \label{fig:logistic_regression}
  \end{subfigure}%
  \begin{subfigure}[b]{0.33\textwidth}
    \centering
    \includegraphics[width=\textwidth]{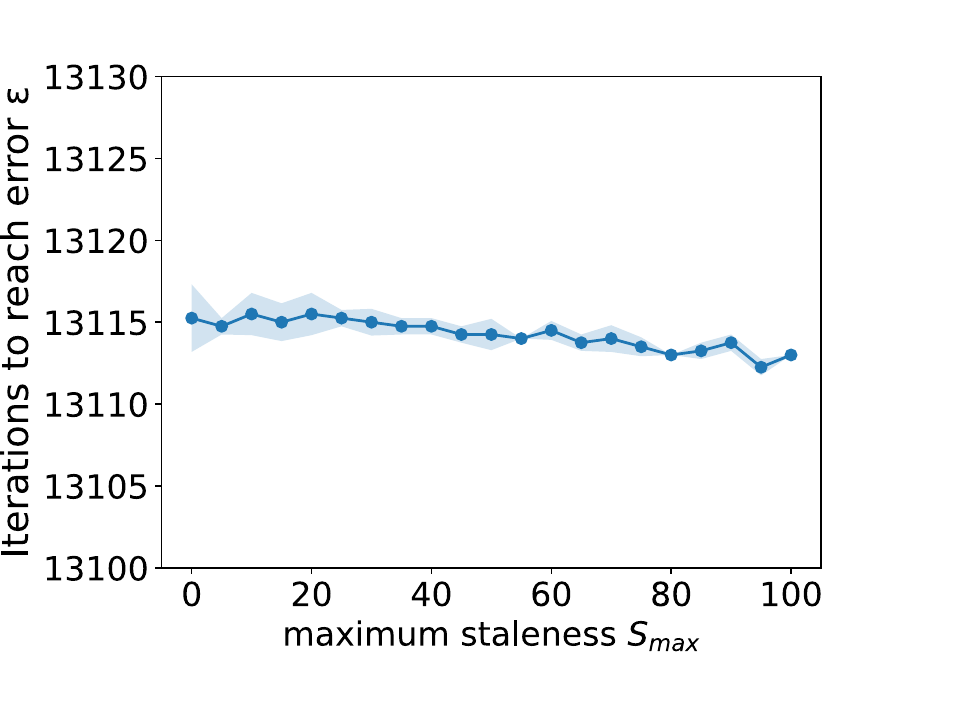}
    \caption{Quadratic function}
    \label{fig:quadratic_function}
  \end{subfigure}
  \begin{subfigure}[b]{0.33\textwidth}
    \centering
    \includegraphics[width=\textwidth]{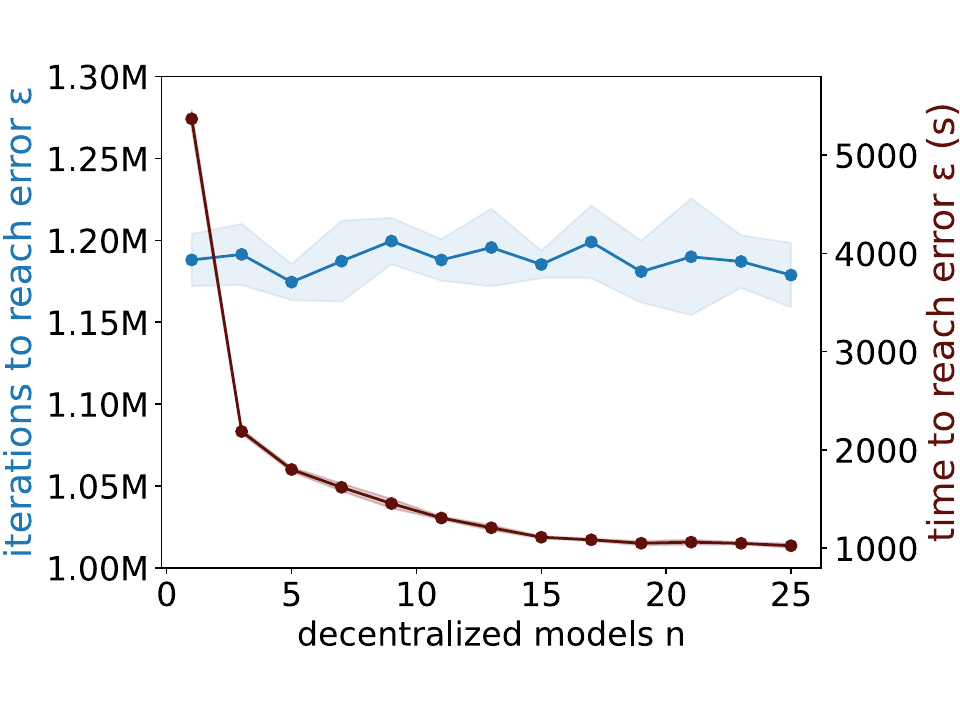}
    \caption{CIFAR}
    \label{fig:cifar}
  \end{subfigure}
  \caption{Numbers of iterations and runtimes needed to reach an error of $\epsilon$ (each averaged over 4 runs, the shaded areas denote one standard deviation).}
  \label{fig:results}
\end{figure}

When analyzing Figure \ref{fig:results}, we observe no significant impact of $\hat{S}_{max}$ on the number of iterations necessary to reach $\epsilon$. This confirms the very good convergence of DASGD in a homogeneous setting with a fixed learning rate, which is largely invariant of $\hat{S}_{max}$ and much better in practice than predicted by our bounds. Consequently, in Figure \ref{fig:cifar}, we observe a considerable decrease in the amount of time necessary to reach $\epsilon$, which exemplifies the practical viability of DASGD.

\section{Conclusion}
In this paper, we prove the convergence rate of decentralized asynchronous SGD (DASGD). Our proof does not depend on partial synchronization among models, complex network topologies, or unrealistic assumptions about the objective function. We introduce a generic staleness measure to quantify the difference between decentralized models over time. Moreover, our resulting bounds show that, over time, the impact of the stochastic noise $\sigma$ on the convergence of the function is higher than the impact of the staleness components $S_{avg}$, $\hat{S}_{max}$, and $\hat{S}_{avg}$, respectively. We demonstrated that our theoretical results are even outperformed by our empirical evaluation over various optimization objectives, including a logistic regression, a quadratic function, and a CNN.


\bibliographystyle{IEEEtran}
\bibliography{bibliography}

\begin{thebibliography}{10}
\providecommand{\url}[1]{#1}
\csname url@samestyle\endcsname
\providecommand{\newblock}{\relax}
\providecommand{\bibinfo}[2]{#2}
\providecommand{\BIBentrySTDinterwordspacing}{\spaceskip=0pt\relax}
\providecommand{\BIBentryALTinterwordstretchfactor}{4}
\providecommand{\BIBentryALTinterwordspacing}{\spaceskip=\fontdimen2\font plus
\BIBentryALTinterwordstretchfactor\fontdimen3\font minus
  \fontdimen4\font\relax}
\providecommand{\BIBforeignlanguage}[2]{{%
\expandafter\ifx\csname l@#1\endcsname\relax
\typeout{** WARNING: IEEEtran.bst: No hyphenation pattern has been}%
\typeout{** loaded for the language `#1'. Using the pattern for}%
\typeout{** the default language instead.}%
\else
\language=\csname l@#1\endcsname
\fi
#2}}
\providecommand{\BIBdecl}{\relax}
\BIBdecl

\bibitem{robbins1951stochastic}
H.~Robbins and S.~Monro, ``A stochastic approximation method,'' \emph{The
  annals of mathematical statistics}, pp. 400--407, 1951.

\bibitem{lecun2002efficient}
Y.~LeCun, L.~Bottou, G.~B. Orr, and K.-R. M{\"u}ller, ``Efficient backprop,''
  in \emph{Neural networks: Tricks of the trade}, 2002, pp. 9--50.

\bibitem{kingma2014adam}
D.~P. Kingma and J.~Ba, ``Adam: A method for stochastic optimization,''
  \emph{arXiv preprint arXiv:1412.6980}, 2014.

\bibitem{duchi2011adaptive}
J.~Duchi, E.~Hazan, and Y.~Singer, ``Adaptive subgradient methods for online
  learning and stochastic optimization.'' \emph{Journal of machine learning
  research}, vol.~12, no.~7, 2011.

\bibitem{openai2023gpt4}
OpenAI, ``Gpt-4 technical report,'' 2023.

\bibitem{rombach2021highresolution}
R.~Rombach, A.~Blattmann, D.~Lorenz, P.~Esser, and B.~Ommer, ``High-resolution
  image synthesis with latent diffusion models,'' 2021.

\bibitem{chen2023symbolic}
X.~Chen, C.~Liang, D.~Huang, E.~Real, K.~Wang, Y.~Liu, H.~Pham, X.~Dong,
  T.~Luong, C.-J. Hsieh \emph{et~al.}, ``Symbolic discovery of optimization
  algorithms,'' \emph{arXiv preprint arXiv:2302.06675}, 2023.

\bibitem{schafl2022hopular}
B.~Sch{\"a}fl, L.~Gruber, A.~Bitto-Nemling, and S.~Hochreiter, ``Hopular:
  Modern hopfield networks for tabular data,'' \emph{arXiv preprint
  arXiv:2206.00664}, 2022.

\bibitem{zhang2020pushing}
Y.~Zhang, J.~Qin, D.~S. Park, W.~Han, C.-C. Chiu, R.~Pang, Q.~V. Le, and Y.~Wu,
  ``Pushing the limits of semi-supervised learning for automatic speech
  recognition,'' \emph{arXiv preprint arXiv:2010.10504}, 2020.

\bibitem{cheng2018model}
Y.~Cheng, D.~Wang, P.~Zhou, and T.~Zhang, ``Model compression and acceleration
  for deep neural networks: The principles, progress, and challenges,''
  \emph{IEEE Signal Processing Magazine}, vol.~35, no.~1, pp. 126--136, 2018.

\bibitem{ouyang2020communication}
S.~Ouyang, D.~Dong, Y.~Xu, and L.~Xiao, ``Communication optimization strategies
  for distributed deep neural network training: A survey,'' \emph{Journal of
  Parallel and Distributed Computing}, vol. 149, pp. 52--65, 2021.

\bibitem{mayer2020scalable}
R.~Mayer and H.-A. Jacobsen, ``Scalable deep learning on distributed
  infrastructures: Challenges, techniques, and tools,'' \emph{ACM Computing
  Surveys (CSUR)}, vol.~53, no.~1, pp. 1--37, 2020.

\bibitem{chen2019round}
C.~Chen, W.~Wang, and B.~Li, ``Round-robin synchronization: Mitigating
  communication bottlenecks in parameter servers,'' in \emph{IEEE INFOCOM
  2019-IEEE Conference on Computer Communications}, 2019, pp. 532--540.

\bibitem{koloskova2022sharper}
A.~Koloskova, S.~U. Stich, and M.~Jaggi, ``{Sharper convergence guarantees for
  asynchronous SGD for distributed and federated learning},'' \emph{Advances in
  Neural Information Processing Systems}, vol.~35, pp. 17\,202--17\,215, 2022.

\bibitem{wu2017decentralized}
T.~Wu, K.~Yuan, Q.~Ling, W.~Yin, and A.~H. Sayed, ``Decentralized consensus
  optimization with asynchrony and delays,'' \emph{IEEE Transactions on Signal
  and Information Processing over Networks}, vol.~4, no.~2, pp. 293--307, 2017.

\bibitem{assran2020asynchronous}
M.~S. Assran and M.~G. Rabbat, ``Asynchronous gradient push,'' \emph{IEEE
  Transactions on Automatic Control}, vol.~66, no.~1, pp. 168--183, 2020.

\bibitem{jiang2021asynchronous}
J.~Jiang, W.~Zhang, J.~Gu, and W.~Zhu, ``Asynchronous decentralized online
  learning,'' \emph{Advances in Neural Information Processing Systems},
  vol.~34, pp. 20\,185--20\,196, 2021.

\bibitem{lian2018asynchronous}
X.~Lian, W.~Zhang, C.~Zhang, and J.~Liu, ``Asynchronous decentralized parallel
  stochastic gradient descent,'' in \emph{International Conference on Machine
  Learning}.\hskip 1em plus 0.5em minus 0.4em\relax PMLR, 2018, pp. 3043--3052.

\bibitem{lian2017can}
X.~Lian, C.~Zhang, H.~Zhang, C.-J. Hsieh, W.~Zhang, and J.~Liu, ``Can
  decentralized algorithms outperform centralized algorithms? a case study for
  decentralized parallel stochastic gradient descent,'' \emph{Advances in
  neural information processing systems}, vol.~30, 2017.

\bibitem{koloskova2019decentralized}
A.~Koloskova, S.~Stich, and M.~Jaggi, ``Decentralized stochastic optimization
  and gossip algorithms with compressed communication,'' in \emph{International
  Conference on Machine Learning}, 2019, pp. 3478--3487.

\bibitem{koloskova2020unified}
A.~Koloskova, N.~Loizou, S.~Boreiri, M.~Jaggi, and S.~Stich, ``{A unified
  theory of decentralized SGD with changing topology and local updates},'' in
  \emph{International Conference on Machine Learning}, 2020, pp. 5381--5393.

\bibitem{ben2019demystifying}
T.~Ben-Nun and T.~Hoefler, ``Demystifying parallel and distributed deep
  learning: An in-depth concurrency analysis,'' \emph{ACM Computing Surveys
  (CSUR)}, vol.~52, no.~4, pp. 1--43, 2019.

\bibitem{recht2011hogwild}
B.~Recht, C.~Re, S.~Wright, and F.~Niu, ``Hogwild!: A lock-free approach to
  parallelizing stochastic gradient descent,'' \emph{Advances in neural
  information processing systems}, vol.~24, 2011.

\bibitem{mishchenko2022asynchronous}
K.~Mishchenko, F.~Bach, M.~Even, and B.~E. Woodworth, ``{Asynchronous SGD beats
  minibatch SGD under arbitrary delays},'' \emph{Advances in Neural Information
  Processing Systems}, vol.~35, pp. 420--433, 2022.

\bibitem{cohen2021asynchronous}
A.~Cohen, A.~Daniely, Y.~Drori, T.~Koren, and M.~Schain, ``Asynchronous
  stochastic optimization robust to arbitrary delays,'' \emph{Advances in
  Neural Information Processing Systems}, vol.~34, pp. 9024--9035, 2021.

\bibitem{arjevani2020tight}
Y.~Arjevani, O.~Shamir, and N.~Srebro, ``{A tight convergence analysis for
  stochastic gradient descent with delayed updates},'' in \emph{Algorithmic
  Learning Theory}, 2020, pp. 111--132.

\bibitem{xie2019asynchronous}
C.~Xie, S.~Koyejo, and I.~Gupta, ``Asynchronous federated optimization,''
  \emph{arXiv preprint arXiv:1903.03934}, 2019.

\bibitem{nguyen2022federated}
J.~Nguyen, K.~Malik, H.~Zhan, A.~Yousefpour, M.~Rabbat, M.~Malek, and D.~Huba,
  ``Federated learning with buffered asynchronous aggregation,'' in
  \emph{International Conference on Artificial Intelligence and Statistics},
  2022, pp. 3581--3607.

\bibitem{toghani2022unbounded}
M.~T. Toghani and C.~A. Uribe, ``Unbounded gradients in federated learning with
  buffered asynchronous aggregation,'' in \emph{2022 58th Annual Allerton
  Conference on Communication, Control, and Computing (Allerton)}, 2022, pp.
  1--8.

\bibitem{stich2020error}
S.~U. Stich and S.~P. Karimireddy, ``{The error-feedback framework: Better
  rates for SGD with delayed gradients and compressed updates},'' \emph{The
  Journal of Machine Learning Research}, vol.~21, no.~1, pp. 9613--9648, 2020.

\bibitem{stich2018local}
\BIBentryALTinterwordspacing
S.~U. Stich, ``Local {SGD} converges fast and communicates little,'' in
  \emph{International Conference on Learning Representations}, 2019. [Online].
  Available: \url{https://openreview.net/forum?id=S1g2JnRcFX}
\BIBentrySTDinterwordspacing

\bibitem{wu2023delay}
X.~Wu, C.~Liu, S.~Magnusson, and M.~Johansson, ``Delay-agnostic asynchronous
  distributed optimization,'' \emph{arXiv preprint arXiv:2303.18034}, 2023.

\bibitem{ram2010asynchronous}
S.~S. Ram, A.~Nedi{\'c}, and V.~V. Veeravalli, ``Asynchronous gossip algorithm
  for stochastic optimization: Constant stepsize analysis,'' in \emph{Recent
  Advances in Optimization and its Applications in Engineering: The 14th
  Belgian-French-German Conference on Optimization}, 2010, pp. 51--60.

\bibitem{sirb2016consensus}
B.~Sirb and X.~Ye, ``Consensus optimization with delayed and stochastic
  gradients on decentralized networks,'' in \emph{2016 IEEE International
  Conference on Big Data (Big Data)}, 2016, pp. 76--85.

\bibitem{even2021asynchronous}
M.~Even, H.~Hendrikx, and L.~Massouli{\'e}, ``Asynchronous speedup in
  decentralized optimization,'' \emph{arXiv preprint arXiv:2106.03585}, 2021.

\bibitem{srivastava2011distributed}
K.~Srivastava and A.~Nedic, ``Distributed asynchronous constrained stochastic
  optimization,'' \emph{IEEE journal of selected topics in signal processing},
  vol.~5, no.~4, pp. 772--790, 2011.

\bibitem{ram2009asynchronous}
S.~S. Ram, A.~Nedi{\'c}, and V.~V. Veeravalli, ``Asynchronous gossip algorithms
  for stochastic optimization,'' in \emph{Proceedings of the 48h IEEE
  Conference on Decision and Control (CDC) held jointly with 2009 28th Chinese
  Control Conference}, 2009, pp. 3581--3586.

\bibitem{luo2020prague}
Q.~Luo, J.~He, Y.~Zhuo, and X.~Qian, ``Prague: High-performance
  heterogeneity-aware asynchronous decentralized training,'' in
  \emph{Proceedings of the Twenty-Fifth International Conference on
  Architectural Support for Programming Languages and Operating Systems}, 2020,
  pp. 401--416.

\bibitem{miao2021heterogeneity}
X.~Miao, X.~Nie, Y.~Shao, Z.~Yang, J.~Jiang, L.~Ma, and B.~Cui,
  ``Heterogeneity-aware distributed machine learning training via partial
  reduce,'' in \emph{Proceedings of the 2021 International Conference on
  Management of Data}, 2021, pp. 2262--2270.

\bibitem{DBLP:conf/iclr/BornsteinRWBH23}
\BIBentryALTinterwordspacing
M.~Bornstein, T.~Rabbani, E.~Wang, A.~S. Bedi, and F.~Huang, ``{SWIFT:} rapid
  decentralized federated learning via wait-free model communication,'' in
  \emph{The Eleventh International Conference on Learning Representations,
  {ICLR} 2023, Kigali, Rwanda, May 1-5, 2023}.\hskip 1em plus 0.5em minus
  0.4em\relax OpenReview.net, 2023. [Online]. Available:
  \url{https://openreview.net/pdf?id=jh1nCir1R3d}
\BIBentrySTDinterwordspacing

\bibitem{agarwal2011distributed}
A.~Agarwal and J.~C. Duchi, ``Distributed delayed stochastic optimization,''
  \emph{Advances in neural information processing systems}, vol.~24, 2011.

\bibitem{feyzmahdavian2016asynchronous}
H.~R. Feyzmahdavian, A.~Aytekin, and M.~Johansson, ``An asynchronous mini-batch
  algorithm for regularized stochastic optimization,'' \emph{IEEE Transactions
  on Automatic Control}, vol.~61, no.~12, pp. 3740--3754, 2016.

\bibitem{lian2015asynchronous}
X.~Lian, Y.~Huang, Y.~Li, and J.~Liu, ``Asynchronous parallel stochastic
  gradient for nonconvex optimization,'' \emph{Advances in neural information
  processing systems}, vol.~28, 2015.

\bibitem{sra2016adadelay}
S.~Sra, A.~W. Yu, M.~Li, and A.~Smola, ``Adadelay: Delay adaptive distributed
  stochastic optimization,'' in \emph{Artificial Intelligence and Statistics},
  2016, pp. 957--965.

\bibitem{dean2012large}
J.~Dean, G.~Corrado, R.~Monga, K.~Chen, M.~Devin, M.~Mao, M.~Ranzato,
  A.~Senior, P.~Tucker, K.~Yang \emph{et~al.}, ``Large scale distributed deep
  networks,'' \emph{Advances in neural information processing systems},
  vol.~25, 2012.

\bibitem{tosi2022convergence}
M.~D. Tosi, V.~Ellampallil~Venugopal, and M.~Theobald, ``Convergence time
  analysis of asynchronous distributed artificial neural networks,'' in
  \emph{9th ACM IKDD CODS and 27th COMAD}, 2022, pp. 314--315.

\bibitem{cnn}
\BIBentryALTinterwordspacing
``{Convolutional Neural Network (CNN): Tensorflow Core},'' 2022, accessed on
  May 10, 2022. [Online]. Available:
  \url{https://www.tensorflow.org/tutorials/images/cnn}
\BIBentrySTDinterwordspacing

\bibitem{krizhevsky2009learning}
A.~Krizhevsky, G.~Hinton \emph{et~al.}, ``Learning multiple layers of features
  from tiny images,'' University of Toronto, Department of Computer Science,
  Tech. Rep., 2009.

\end{thebibliography}

\clearpage

\appendix
\section{Appendix}
\label{sec:Appendix}
\subsection{Useful Inequalities \& Remarks}

\subsubsection{Inequalities}

Below, we list a number of useful inequalities to which we will refer in our proof of Theorem \ref{theorem}.

\smallskip
\begin{lemma}
\label{plemma1}
    In analogy to \cite{koloskova2022sharper}, we establish the following inequality for any set of $n$ vectors $\{a_i\}^{n}_{i=1}$ with $a_i \in \mathbb{R}^d$:
    \begin{equation}
    \label{inequality1}
        \|\sum^{n}_{i=1}a_i\|^2 \leq n \sum^{n}_{i=1}\|a_i\|^2
    \end{equation}
\end{lemma}

\smallskip
\begin{lemma}
\label{plemma2}
For a vector $a \in \mathbb{R}^d$ and a multiplier $m \in \mathbb{R}$, it holds that:
    \begin{equation}
    \label{inequality2}
        \|ma\|^2 \leq m^2\|a\|^2
    \end{equation}
\end{lemma}

\smallskip
\begin{lemma}
\label{plemma3}
Considering Assumption \ref{as:l-smooth}, for any function $f$, it holds that:
    \begin{equation}
    \label{inequality3}
        \| \nabla f(y)\| \leq \|\nabla f(x)\| + L\|x - y\| \hspace*{40pt} \forall x,y \in \mathbb{R}^d
    \end{equation}
\end{lemma}

\smallskip
\begin{lemma}
\label{plemma4}
    From the polarization identity, we have:
    \begin{equation}
    \label{inequality4}
        \langle a,b \rangle = \frac{\|a\|^2}{2} + \frac{\|b\|^2}{2} - \frac{\|a-b\|^2}{2}
    \end{equation}
\end{lemma}

\smallskip
\begin{lemma}
\label{plemma5}
    For any $a, b \in \mathbb{R}$, it holds that:
    \begin{equation}
    \label{inequality5}
        (a+b)^2 \leq 2a^2 + 2b^2
    \end{equation}
\end{lemma}

\subsubsection{Remarks}
Here, we provide the following useful remarks.

\begin{remark}\label{rmk-Deltabound}
$\Delta_{i,j}^{t,s}$ is larger or equal to the sum of any subset $A \subseteq \hat{S}_{i,j}^{t,s}$ (scaled with $\eta$), that is:
\begin{align*}
     \eta \sum_{g \in A} |g| \leq \Delta_{i,j}^{t,s}
\end{align*}
\end{remark}

We can guarantee this because $\Delta_{i,j}^{t,s}$ is the sum of the absolute values of all gradients in $\hat{S}_{i,j}^{t,s}$ (also scaled with $\eta$). Take as an example a set $\hat{S}_{i,j}^{t,s} = \{g_{\alpha}, g_{\beta}, g_{\gamma} \}$. Thus, $\Delta_{i,j}^{t,s} = \eta(|g_{\alpha}| + |g_{\beta}| + |g_{\gamma}|)$. Evidently, $\Delta_{i,j}^{t,s} \geq \eta|g_{\alpha}|$. Consequently, we can also reach the following remark.

\begin{remark}\label{rmk-DeltaUpperbound}
Considering $|x_{\alpha}| = x^t_i - \eta \sum_{g \in A} |g|$ and $A \subseteq \hat{S}_{i,j}^{t,s}$, we can guarantee that:
    \begin{align}
        x^t_i - |x_{\alpha}| \leq \Delta_{i,j}^{t,s}
        \label{eq:rmk-DeltaUpperbound}
    \end{align}
\end{remark}

Remark \ref{rmk-Deltabound} follows the same idea as Remark \ref{rmk-DeltaUpperbound}. By definition, $\Delta_{i,j}^{t,s} = x^t_i - |x_{\alpha}| - |x_{\beta}|$, considering $|x_{\beta}| = x^t_i - \eta \sum_{g \in B} |g|$ and $B$ being the complementary set of $A$ and $\hat{S}_{i,j}^{t,s}$, that is $B = \overline{(A \cup \hat{S}_{i,j}^{t,s})}$. Thus, $x^t_i - |x_{\alpha}| \leq \Delta_{i,j}^{t,s}$.

\subsection{Proof of Theorem \ref{theorem}}
\label{proof}
Our proof for Theorem \ref{theorem} closely follows the structure of the proofs provided in Koloskova, Stich, and Jaggi in \cite{koloskova2022sharper} (specifically the ones leading to Theorem 6).

First, let us recall our Theorem \ref{theorem} from Section \ref{sec:convergence-analysis}.

\begin{customthm}{1}
Considering Assumptions \ref{as:bounded-variance}, \ref{as:func-heter}, \ref{as:l-smooth}, \ref{as:bounded-gradient}, and a constant stepsize $\eta \leq \frac{1}{4LS_{avg}}$, Algorithm \ref{alg:model} reaches $\frac{1}{T+1} \sum^{T}_{t=0}( \| \nabla f(x^{t})\|^2) \leq \epsilon$ after 

\begin{equation}\tag{\ref{eq:1theorem}}
\mathcal{O}(\frac{\sigma}{\epsilon^2}) + \mathcal{O}(\frac{QS_{avg}}{\epsilon^\frac{3}{2}}) + \mathcal{O}(\frac{S_{avg}}{\epsilon}) \hspace*{40pt} \text{iterations}.
\end{equation}

Moreover, without Assumption \ref{as:bounded-gradient} and $\eta \leq \frac{1}{4L\sqrt{\hat{S}_{avg}\hat{S}_{max}}}$, Algorithm \ref{alg:model} reaches $\frac{1}{T+1} \sum^{T}_{t=0}(\|\nabla f(x^{t})\|^2) \leq \epsilon$ after

\begin{equation}\tag{\ref{eq:2theorem}}
\mathcal{O}(\frac{\sigma^2}{\epsilon^2}) + \mathcal{O}(\frac{\sqrt{\hat{S}_{avg}\hat{S}_{max}}}{\epsilon}) \hspace*{60pt} \text{iterations}.
\end{equation}
\end{customthm}

First, we define $\delta_{i,j}^{t,s} \in \mathbb{R}^d$ as the difference between two models $x^t_i$ and $x^s_j$, which can also be represented as the sum of the gradients in the staleness set $S_{i,j}^{t,s}$ scaled by $\eta$, as follows:
\begin{equation}
    \delta_{i,j}^{t,s} = \sum_{g \in S_{i,j}^{t,s}} \eta g
\end{equation}

Then, we represent its upper bound $\Delta_{i,j}^{t,s}, \in \mathbb{R}^d$ as the summation of the absolute values of the gradients in $S_{i,j}^{t,s}$ (also scaled by $\eta$), expressed as follows:

\begin{equation}
    \Delta_{i,j}^{t,s} = \sum_{g \in S_{i,j}^{t,s}} \eta |g| \geq \delta_{i,j}^{t,s}
\end{equation}

Now, we define Lemma \ref{lemma1} which will be used to reach the convergence rates of Equations (\ref{eq:1theorem}) and (\ref{eq:2theorem}).

\bigskip
\begin{lemma}[Descent Lemma]\label{lemma1} Considering Assumptions \ref{as:bounded-variance}, \ref{as:func-heter} and \ref{as:l-smooth} with a stepsize $\eta \leq \frac{1}{2L}$, we have:
\begin{equation}
     \mathbb{E}_{t+1} \left[f(x^{t+1}_i)\right] \leq f(x^{t}_i) - \frac{\eta}{2}\| \nabla f(x^{t}_i)\|^2 + L\eta^2\sigma^2 + \frac{\eta L^2}{2}\|\Delta_{i,j}^{t,s}\|^2
\end{equation}
\end{lemma}

{\em Proof.}

Following \cite{nguyen2022federated, koloskova2022sharper} and due to the L-smoothness of $f$, when model $x^t_i$ updates itself with a gradient computed by model $x^s_j$, it then holds that:

\begin{align*}
     \mathbb{E}_{t+1} \left[f(x^{t+1}_i)\right] &=  \mathbb{E}_{t+1} \left[f(x^{t}_i - \eta \nabla F(x^{t}_i+\delta_{i,j}^{t,s}, \xi^s_j))\right]\\
     &\leq f(x^{(t)}) - \underbrace{\eta~  \mathbb{E}_{t+1}\left[ \langle \nabla f(x^{t}_i), \nabla F(x^{t}_i+\delta_{i,j}^{t,s}, \xi^s_j) \rangle\right]}_{T_1} + ~\mathbb{E}_{t+1}\left[\frac{L\eta^2}{2}\underbrace{\|\nabla F(x^{t}_i+\delta_{i,j}^{t,s}, \xi^s_j)\|^2}_{T_2}\right]
\end{align*}

Note that, due to the function homogeneity assumption (Assumption \ref{as:func-heter}), we omit the indices of $F$ and $f$ throughput our proofs to simplify their notations.

We first transform $T_1$ as follows:

\begin{align*}
    T_1 &= -\eta~ \mathbb{E}_{t+1} \left[ \langle \nabla f(x^{t}_i), \nabla F(x^{t}_i +\delta_{i,j}^{t,s}, \xi^s_j)\rangle\right] \\
    &= -\eta~ \langle \nabla f(x^{t}_i), \nabla f(x^{t}_i +\delta_{i,j}^{t,s})\rangle\\
    &\stackrel{(\ref{inequality4})}{=} - \frac{\eta}{2} \|\nabla f(x^{t}_i)\|^2 - \frac{\eta}{2}\|\nabla f(x^{t}_i + \delta_{i,j}^{t,s})\|^2 + \frac{\eta}{2}\|\nabla f(x^{t}_i) - \nabla f(x^{t}_i + \delta_{i,j}^{t,s})\|^2 
\end{align*}

Next, we transform $T_2$ as follows:

\begin{align*}
    T_2 &= \mathbb{E}_{t+1} \left[\| \nabla F(x^{t}_i + \delta_{i,j}^{t,s}, \xi^s_j)\|^2\right]\\
    &\stackrel{(\ref{eq:bounded-variance})}{\leq} \sigma^2 + \|\nabla f(x^{t}_i + \delta_{i,j}^{t,s})\|^2
\end{align*}

Then, we combine again $T_1$ and $T_2$:

\begin{equation*}
    \mathbb{E}_{t+1} \left[f(x^{t+1}_i)\right] \leq f(x^{t}_i) - \frac{\eta}{2} \|\nabla f(x^{t}_i)\|^2 - \frac{\eta}{2}(1-L\eta)\|\nabla f(x^{t}_i + \delta_{i,j}^{t,s})\|^2 + \frac{\eta}{2}\|\nabla f(x^{t}_i) - \nabla f(x^{t}_i + \delta_{i,j}^{t,s})\|^2 +  \frac{L\eta^2\sigma^2}{2}
\end{equation*}

By exploiting the L-smoothness to estimate $\|\nabla f(x^{t}_i) - \nabla f(x^{t}_i + \delta_{i,j}^{t,s})\|^2 \leq L^2 \|x^{t}_i - (x^{t}_i + \delta_{i,j}^{t,s})\|^2$, we obtain:

\begin{equation*}
    \mathbb{E}_{t+1} \left[f(x^{t+1}_i)\right] \stackrel{(\ref{eq:l-smooth})}{\leq} f(x^{t}_i) - \frac{\eta}{2} \|\nabla f(x^{t}_i)\|^2 - \frac{\eta}{2}(1- L\eta) \| \nabla f(x^{t}_i + \delta_{i,j}^{t,s}) \|^2 + \frac{\eta L^2}{2}\|x^{t}_i - (x^{t}_i + \delta_{i,j}^{t,s})\|^2 + L\eta^2\sigma^2
\end{equation*}

Simplifying the fourth term, we get: 

\begin{equation*}
\mathbb{E}_{t+1} \left[f(x^{t+1}_i)\right] \leq f(x^{t}_i) - \frac{\eta}{2} \|\nabla f(x^{t}_i)\|^2 - \frac{\eta}{2}(1- L\eta) \| \nabla f(x^{t}_i + \delta_{i,j}^{t,s}) \|^2 + \frac{\eta L^2}{2}\| \delta_{i,j}^{t,s}\|^2 + L\eta^2\sigma^2
\end{equation*}

By applying $\eta \leq \frac{1}{2L}$, we then obtain:

\begin{align*}
     \leq f(x^{t}_i) - \frac{\eta}{2}\| \nabla f(x^{t}_i)\|^2 - \frac{\eta}{4}\|\nabla f(x^{t}_i+\delta_{i,j}^{t,s})\|^2 + L\eta^2\sigma^2 + \frac{\eta L^2}{2}\|\delta_{i,j}^{t,s}\|^2
\end{align*}

By discarding $- \frac{\eta}{4}\|\nabla f(x^{t}_i+\delta_{i,j}^{t,s})\|^2$ from the right-hand side of the inequality and considering $\Delta_{i,j}^{t,s} \geq \delta_{i,j}^{t,s}$, we reach Lemma \ref{lemma1}.\qed

\subsubsection{Preliminaries for the Proof of Theorem \ref{theorem} with the Convergence Rate of Equation (\ref{eq:1theorem})}

\begin{lemma}[Estimation of the residual -- bounded gradients]\label{lemma3}
Considering Assumptions \ref{as:bounded-variance}, \ref{as:func-heter}, \ref{as:l-smooth} and \ref{as:bounded-gradient} with a constant stepsize $\eta \leq \frac{1}{4LS_{avg}}$, we have:

\begin{equation*}
     \frac{1}{T+1}\sum^T_{i=0}~\mathbb{E} \left[\|\Delta_{i,j}^{t,s}\|^2 \right]\leq S_{avg}^2\eta^2Q^2 + S_{avg} \eta^2 \sigma^2
\end{equation*}

\end{lemma}

{\em Proof.}

From Equation (\ref{eq:boundedgradient-proof}), when using the tighter representation of staleness $|S_{i,j}^{t,s}|$, we obtain:

\begin{equation*}
    \leq |S_{i,j}^{t,s}|\eta^2~\mathbb{E} \left[\sum_{g^u_k \in S_{i,j}^{t,s}}\|\nabla f(x^u_k)\|^2 \right] + |S_{i,j}^{t,s}|\eta^2\sigma^2
\end{equation*}

By considering Assumption \ref{as:bounded-gradient}, in which $\|\nabla f(x^u_k)\|^2 \leq Q^2$, we have:

\begin{align*}
     ~\mathbb{E} \left[\|\Delta_{i,j}^{t,s}\|^2 \right] \stackrel{(\ref{eq:bounded-gradient})}{\leq} |S_{i,j}^{t,s}|^2\eta^2Q^2 + |S_{i,j}^{t,s}| \eta^2 \sigma^2
\end{align*}

Then, by averaging over all steps $t \in T$, we already obtain the statement of the lemma. \qed

\subsubsection{Proof of Theorem \ref{theorem} with the Convergence Rate of Equation (\ref{eq:1theorem})}

We are now ready to give the proof of Theorem \ref{theorem} with the convergence rate of Equation (\ref{eq:1theorem}). We start by passing $f(x^{t+1}_i)$ to the right-hand side of the inequality of Lemma \ref{lemma1} and $f(x^{t}_i)$ to the left-hand side, respectively.

\begin{align*}
    \frac{\eta}{2}\| \nabla f(x^{t}_i)\|^2   \leq f(x^{t}_i) - ~\mathbb{E}_{t+1} \left[f(x^{t+1}_i) \right] + L\eta^2\sigma^2 + \frac{\eta L^2}{2}\|\Delta_{i,j}^{t,s}\|^2
\end{align*}

Then, we average over all steps $t \in T$ and divide by $\eta$.

\begin{align*}
    \frac{1}{T+1} \sum^{T}_{t=0}\frac{1}{2} ~\mathbb{E} \left[ \| \nabla f(x^{t}_i)\|^2 \right] \leq \frac{1}{\eta(T+1)}(f(x^{0}_i - f^*) + L\eta \sigma^2 + \frac{1}{T+1}\frac{L^2}{2}\sum^{T}_{t=0} ~\mathbb{E} \left[ \| \Delta_{i,j}^{t,s}\|^2 \right]
\end{align*}

We next apply Lemma 3 to the last term.

\begin{align*}
    \frac{1}{T+1} \sum^{T}_{t=0}\frac{1}{2} ~\mathbb{E} \left[ \| \nabla f(x^{t}_i)\|^2 \right] \leq \frac{1}{\eta(T+1)}(f(x^{0}_i - f^*) + L\eta \sigma^2 + \frac{L^2S_{avg}^2\eta^2Q^2}{2} + \frac{L^2S_{avg} \eta^2 \sigma^2}{2}
\end{align*}

Let again $f(x^{0}_i) - f^* =: r_0$. We simplify the inequality with the following conditions:
\begin{itemize}
    \item we multiply the inequality by 2;
    \item let $\eta \leq \frac{1}{4LS_{avg}}$ on the last term of the inequality;
    \item we consider that $\frac{2L^2S_{avg} \eta^2 \sigma^2}{2} \leq \frac{L \eta \sigma^2}{4} \leq L \eta \sigma^2$.
\end{itemize}

We therefore obtain:

\begin{align*}
    \frac{1}{T+1} \sum^{T}_{t=0}~\mathbb{E} \left[\| \nabla f(x^{t}_i)\|^2 \right] \leq \frac{2r_0}{\eta(T+1)} + 3L\eta \sigma^2 + L^2 S_{avg}^2\eta^2Q^2
\end{align*}

Let us assume $\eta \leq \frac{1}{4LS_{avg}}$, then together with Lemma 17 from \cite{koloskova2020unified}, we obtain:

\begin{align*}
    \Psi_T\leq 2(\frac{3L\sigma^2r_0}{T+1})^{\frac{1}{2}} + 2(L^2S_{avg}^2Q^2)^{\frac{1}{3}}(\frac{r_0}{T+1})^\frac{2}{3}+\frac{4Lr_0S_{avg}}{T+1}
\end{align*}

This finally yields the bound of Theorem \ref{theorem} with the convergence rate of Equation (\ref{eq:1theorem}).

\begin{align*}
    \mathcal{O}(\frac{\sigma}{\sqrt{T}}) + \mathcal{O}(\frac{QS_{avg}}{T^\frac{2}{3}}) + \mathcal{O}(\frac{S_{avg}}{T})
\end{align*}\qed

\subsubsection{Preliminaries for the Proof of Theorem \ref{theorem} with the Convergence Rate of Equation (\ref{eq:2theorem})}

\bigskip
\begin{lemma}[Estimation of the residual]\label{lemma2}
By considering Assumptions \ref{as:bounded-variance}, \ref{as:func-heter} \ref{as:l-smooth} and a constant stepsize $\eta \leq \frac{1}{4L\sqrt{\hat{S}_{avg}\hat{S}_{max}}}$, we have:

\begin{equation*}
    \frac{1}{(T+1)}\sum^T_{t=0} ~\mathbb{E} \left[\|\Delta_{i,j}^{t,s}\|^2 \right] \leq \frac{1}{7L^2(T+1)}\sum^T_{t=0} ~\mathbb{E} \left[\|\nabla f(x_{i}^{t})\|^2 \right] + \frac{2\eta\sigma^2}{7L}
\end{equation*}
\end{lemma}

{\em Proof.}

First, we unroll $\Delta_{i,j}^{t,s}$ as follows:
\begin{align*}
    ~\mathbb{E}  \left[\| \Delta_{i,j}^{t,s}\|^2 \right]  &= ~\mathbb{E} \left[ \left\| \sum_{g \in \hat{S}_{i,j}^{t,s}} \eta |g| \right\|^2 \right] 
\end{align*}

Let $g^u_k \in \hat{S}_{i,j}^{t,s}$ be a gradient calculated by any model $k$ at any iteration $u$, then we have:
\begin{align*}
    = ~\mathbb{E} \left[\left\|\sum_{g^u_k \in \hat{S}_{i,j}^{t,s}} \eta |\nabla F(x^u_k, \xi^u_k)|\right\|^2 \right]
\end{align*}

Now, by considering Lemmas \ref{plemma1} and \ref{plemma2} as well as Assumption \ref{as:bounded-variance}, we have:

\begin{align}
    &\stackrel{(\ref{eq:bounded-variance})}{\leq} ~\mathbb{E} \left[\left\|\sum_{g^u_k \in \hat{S}_{i,j}^{t,s}} \eta |\nabla f(x^u_k)|\right\|^2 \right] + |\hat{S}_{i,j}^{t,s}|\eta^2\sigma^2 \\
    &\stackrel{(\ref{inequality1})}{\leq} 
    |\hat{S}_{i,j}^{t,s}| ~\mathbb{E}\left[\sum_{g^u_k \in \hat{S}_{i,j}^{t,s}} \eta \| \nabla f(x^u_k)\|^2\right] + |\hat{S}_{i,j}^{t,s}|\eta^2\sigma^2 \\
    &\stackrel{(\ref{inequality2})}{\leq} 
    \label{eq:boundedgradient-proof}|\hat{S}_{i,j}^{t,s}| \eta^2~\mathbb{E}\left[\sum_{g^u_k \in \hat{S}_{i,j}^{t,s}} \| \nabla f(x^u_k)\|^2\right] + |\hat{S}_{i,j}^{t,s}|\eta^2\sigma^2
\end{align}

By considering the L-smoothness to estimate $\nabla f(x^u_k)$, we obtain:

\begin{align*}
    \stackrel{(\ref{inequality3})}{\leq} |\hat{S}_{i,j}^{t,s}|\eta^2~\mathbb{E} \left[\sum_{g^u_k \in \hat{S}_{i,j}^{t,s}} (\|\nabla f(x_{i}^{t})\| + L\|x^t_i - x^u_k\|)^2 \right] + |\hat{S}_{i,j}^{t,s}|\eta^2\sigma^2
\end{align*}

Next, consider Remark \ref{rmk-DeltaUpperbound}.

\begin{align*}
    \stackrel{(\ref{eq:rmk-DeltaUpperbound})}{\leq}  |\hat{S}_{i,j}^{t,s}|\eta^2~\mathbb{E} \left[\sum_{g \in \hat{S}_{i,j}^{t,s}} (\|\nabla f(x_{i}^{t})\|+L\|\Delta^{t,s}_{i,j}\|)^2 \right]+ |\hat{S}_{i,j}^{t,s}|\eta^2\sigma^2
\end{align*}

Let us assume $\eta \leq \frac{1}{4L\sqrt{\hat{S}_{avg}\hat{S}_{max}}}$ in the first term of the right-hand side of the inequality and $|\hat{S}_{i,j}^{t,s}| \leq \hat{S}_{max}$. We then obtain:

\begin{align*}
    \leq \frac{1}{16L^2 \hat{S}_{avg}}~\mathbb{E} \left[\sum_{g \in \hat{S}_{i,j}^{t,s}} (\|\nabla f(x_{i}^{t})\|+L\|\Delta^{t,s}_{i,j}\|)^2 \right]+ |\hat{S}_{i,j}^{t,s}|\eta^2\sigma^2
\end{align*}

By summing over all steps $t \in T$, we obtain:

\begin{align*}
    \sum^T_{t=0} ~\mathbb{E} \left[\|\Delta_{i,j}^{t,s}\|^2  \right]\stackrel{(\ref{eq:s_avg*})}{\leq} \frac{1}{16L^2\hat{S}_{avg}} \sum^T_{t=0} \sum_{g \in \hat{S}_{i,j}^{t,s}} ~\mathbb{E} \left[\|\nabla f(x_{i}^{t})\|+L\|\Delta^{t,s}_{i,j}\|\right]^2 + (T+1) \hat{S}_{avg}\eta^2\sigma^2
\end{align*}
  
By considering that the inner summation from the right side of the equation is bounded by $|\hat{S}_{i,j}^{t,s}|$ and is executed $T$ times, we can simplify the equation by using Definition \ref{def:s_avg}. This simplification upper bounds the number of iterations over $\nabla f(x_{i}^{t})$ and $\Delta^{t,s}_{i,j}$ to $\hat{S}_{avg}$ times, thus leading us to:

\begin{align*}
    \stackrel{(\ref{eq:s_avg*})}{\leq} \frac{1}{16L^2}\sum^T_{t=0} ~\mathbb{E}\left[\|\nabla f(x_{i}^{t})\|+L\|\Delta^{t,s}_{i,j}\|\right]^2 + (T+1)\hat{S}_{avg}\eta^2\sigma^2
\end{align*}

From Lemma \ref{plemma5}, we consider that  $~\mathbb{E}\left[\|\nabla f(x_{i}^{t})\|+L\|\Delta^{t,s}_{i,j}\|\right]^2 \leq ~\mathbb{E}\left[ 2\|\nabla f(x_{i}^{t})\|^2+2L^2\|\Delta^{t,s}_{i,j}\|^2\right]$, thus:

\begin{align*}
    \sum^T_{t=0} ~\mathbb{E} \left[\|\Delta_{i,j}^{t,s}\|^2  \right] \leq \frac{1}{16L^2}\sum^T_{t=0} ~\mathbb{E}\left[ 2\|\nabla f(x_{i}^{t})\|^2+2L^2\|\Delta^{t,s}_{i,j}\|^2\right] + (T+1)\hat{S}_{avg}\eta^2\sigma^2
\end{align*}

Then, by taking all terms based on $\|\Delta^{t,s}_{i,j}\|$ to the left side of the inequality:

\begin{align*}
    (1-\frac{2}{16}) \sum^T_{t=0} ~\mathbb{E} \left[\|\Delta_{i,j}^{t,s}\|^2  \right] \leq \frac{1}{16L^2}\sum^T_{t=0} ~\mathbb{E}\left[ 2\|\nabla f(x_{i}^{t})\|^2\right] + (T+1)\hat{S}_{avg}\eta^2\sigma^2
\end{align*}

Later simplifying the inequality, we reach:

\begin{align*}
    \frac{7}{8} \sum^T_{t=0} ~\mathbb{E} \left[\|\Delta_{i,j}^{t,s}\|^2  \right] \leq \frac{1}{8L^2}\sum^T_{t=0} ~\mathbb{E}\left[ \|\nabla f(x_{i}^{t})\|^2\right] + (T+1)\hat{S}_{avg}\eta^2\sigma^2
\end{align*}

Then, by multiplying the inequality by $8/7$, we have:

\begin{align*}
    \sum^T_{t=0} ~\mathbb{E} \left[\|\Delta_{i,j}^{t,s}\|^2  \right] \leq \frac{1}{7L^2}\sum^T_{t=0} ~\mathbb{E}\left[ \|\nabla f(x_{i}^{t})\|^2\right] + \frac{8(T+1)}{7}\hat{S}_{avg}\eta^2\sigma^2
\end{align*}

At last, by assuming $\eta \leq \frac{1}{4L\sqrt{\hat{S}_{max}\hat{S}_{avg}}}$, and $\hat{S}_{max} \geq \hat{S}_{avg}$ such that $\sqrt{\hat{S}_{max}S_{avg}} \geq \hat{S}_{avg}$. Then, it holds that:

\begin{align*}
    \sum^T_{t=0} ~\mathbb{E} \left[\|\Delta_{i,j}^{t,s}\|^2  \right] \leq \frac{1}{7L^2}\sum^T_{t=0} ~\mathbb{E}\left[ \|\nabla f(x_{i}^{t})\|^2\right] + \frac{2(T+1)}{7}L\eta\sigma^2
\end{align*}

After dividing by $(T+1)$, we reach the statement of the lemma.\qed

\subsubsection{Proof of Theorem \ref{theorem} with the Convergence Rate of Equation (\ref{eq:2theorem})}

Finally, we give the proof of Theorem \ref{theorem} with the convergence rate of Equation (\ref{eq:2theorem}). We start by passing $f(x^{t+1}_i)$ to the right-hand side of the inequality of Lemma \ref{lemma1} and $f(x^{t}_i)$ to the left-hand side. Then, for every model $i \in [n]$, it holds that:

\begin{align*}
    \frac{\eta}{2}\| \nabla f(x^{t}_i)\|^2 \leq f(x^{t}_i) - ~\mathbb{E}_{t+1} \left[ f(x^{t+1}_i) \right] + L\eta^2\sigma^2 + \frac{\eta L^2}{2}\|\Delta_{i,j}^{t,s}\|^2
\end{align*}

Then, we average over all $t \in T$ and divide by $\eta$. In the following, let $f*$ denote the value of our objective function at a local minimum $\epsilon$.

\begin{align*}
    \frac{1}{T+1} \sum^{T}_{t=0}\frac{1}{2} ~\mathbb{E} \left[ \| \nabla f(x^{t}_i)\|^2 \right] \leq \frac{1}{\eta(T+1)}(f(x^{0}_i) - f^*) + L\eta \sigma^2 + \frac{1}{(T+1)}\frac{L^2}{2}\sum^{T}_{t=0} ~\mathbb{E} \left[ \| \Delta_{i,j}^{t,s}\|^2 \right]
\end{align*}

We next apply Lemma \ref{lemma2} to the last term in order to obtain:

\begin{align*}
    \frac{1}{T+1} \sum^{T}_{t=0}\frac{1}{2} ~\mathbb{E} \left[\| \nabla f(x^{t}_i)\|^2 \right] \leq \frac{1}{\eta(T+1)}(f(x^{0}_i) - f^*) + L\eta \sigma^2 + \frac{1}{14(T+1)}\sum^{T}_{t=0} ~\mathbb{E} \left[ \| \nabla f(x^{t}_i)\|^2 \right] + \frac{L\eta\sigma^2}{7}
\end{align*}

By setting $f(x^{0}_i) - f^* =: r_0$ and considering that $\frac{L\eta\sigma^2}{7} < L\eta\sigma^2$, we get:

\begin{align*}
    \frac{1}{T+1} \sum^{T}_{t=0}(\frac{1}{2} ~\mathbb{E} \left[ \| \nabla f(x^{t}_i)\|^2) \right] \leq \frac{r_0}{\eta(T+1)} + 2L\eta \sigma^2+ \frac{1}{14(T+1)}\sum^{T}_{t=0} ~\mathbb{E} \left[\| \nabla f(x^{t}_i)\|^2 \right]
\end{align*}

We next pass $\mathbb{E} \left[\| \nabla f(x^{t}_i)\|^2 \right]$ to the left-hand side of the inequality.

\begin{align*}
    \frac{1}{T+1} \sum^{T}_{t=0}(\frac{1}{2} - \frac{1}{14}) ~\mathbb{E} \left[\| \nabla f(x^{t}_i)\|^2 \right] \leq \frac{r_0}{\eta(T+1)} + 2L\eta \sigma^2\\
    \frac{1}{T+1} \sum^{T}_{t=0}\frac{3}{7}~\mathbb{E} \left[\| \nabla f(x^{t}_i)\|^2 \right] \leq \frac{r_0}{\eta(T+1)} + 2L\eta \sigma^2
\end{align*}

We then multiply the inequality by $\frac{7}{3}$ as follows:

\begin{equation*}
    \frac{1}{T+1} \sum^{T}_{t=0}~\mathbb{E} \left[ \| \nabla f(x^{t}_i)\|^2 \right] \leq \frac{7r_0}{3\eta(T+1)} + \frac{14}{3}L\eta \sigma^2
\end{equation*}

Using $\eta \leq \frac{1}{4L\sqrt{\hat{S}_{avg}\hat{S}_{max}}}$ together with Lemma 17 from \cite{koloskova2020unified}.

\begin{align*}c
    \Psi_T\leq 2(\frac{14L\sigma^2r_0}{3(T+1)})^{\frac{1}{2}} +\frac{4L r_0\sqrt{\hat{S}_{avg}\hat{S}_{max}}}{T+1}
\end{align*}

This finally yields the bound of Theorem \ref{theorem} with the convergence rate of Equation (\ref{eq:2theorem}).

\begin{align*}
    \mathcal{O}(\frac{\sigma}{\sqrt{T}}) + \mathcal{O}(\frac{\sqrt{\hat{S}_{avg} \hat{S}_{max}}}{T})
\end{align*}\qed

\end{document}